\documentclass[10pt,twocolumn,letterpaper]{article}

\usepackage{cvpr}
\usepackage{times}
\usepackage{epsfig}
\usepackage{graphicx}
\usepackage{amsmath}
\usepackage{amssymb}
\usepackage[pagebackref=true,colorlinks=true,breaklinks=true,bookmarks=false,citecolor=blue]{hyperref}

\cvprfinalcopy

\def\cvprPaperID{8915}

\ifcvprfinal\fi

\begin{document}
%%%%%%%%%% TITLE
\title{STEFANN: Scene Text Editor using Font Adaptive Neural Network}
\author{
  Prasun Roy$^{1}$\thanks{These authors contributed equally to this work.}~,~~
  Saumik Bhattacharya$^{2}$\footnotemark[1]~,~~
  Subhankar Ghosh$^{1}$\footnotemark[1]~,~~
  and Umapada Pal$^{1}$ \\
  $^{1}$Indian Statistical Institute, Kolkata, India \\
  $^{2}$Indian Institute of Technology, Kharagpur, India \\
  % {\tt\small \{prasunroy.pr, soumiksweb, sgcs2005\}@gmail.com, umapada@isical.ac.in} \\
  {\tt\small \url{https://prasunroy.github.io/stefann}} \\
}
\maketitle
\thispagestyle{empty}

%%%%%%%%%% CHAPTERS
\begin{abstract}
Textual information in a captured scene plays an important role in scene interpretation and decision making. Though there exist methods that can successfully detect and interpret complex text regions present in a scene, to the best of our knowledge, there is no significant prior work that aims to modify the textual information in an image. The ability to edit text directly on images has several advantages including error correction, text restoration and image reusability.
In this paper, we propose a method to modify text in an image at character-level. We approach the problem in two stages. At first, the unobserved character (target) is generated from an observed character (source) being modified. We propose two different neural network architectures -- (a) \textbf{FANnet} to achieve structural consistency with source font and (b) \textbf{Colornet} to preserve source color. Next, we replace the source character with the generated character maintaining both geometric and visual consistency with neighboring characters. Our method works as a unified platform for modifying text in images. We present the effectiveness of our method on COCO-Text and ICDAR datasets both qualitatively and quantitatively.
\end{abstract}

\section{Introduction}
\begin{figure}
\centering
\includegraphics[width=\linewidth]{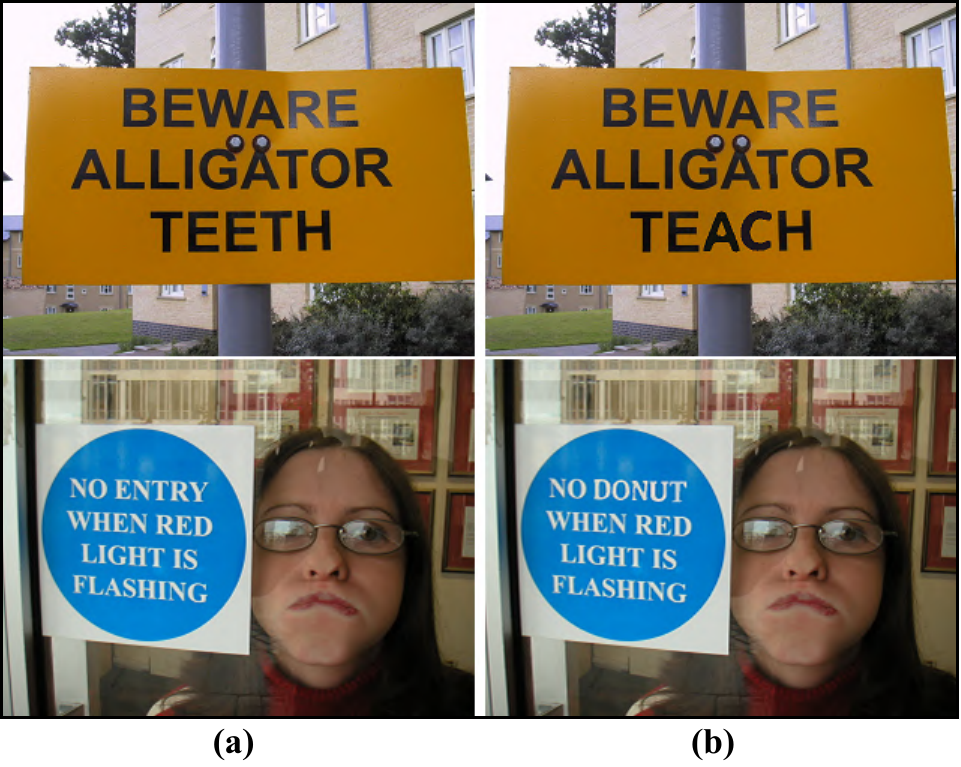}
\caption{Examples of text editing using STEFANN: \textbf{(a)} Original images from ICDAR dataset; \textbf{(b)} Edited images. It can be observed that STEFANN can edit multiple characters in a word (top row) as well as an entire word (bottom row) in a text region.}
\label{fig:introduction}
\end{figure}

Text is widely present in different design and scene images. It contains important contextual information for the readers. However, if any alteration is required in the text present in an image, it becomes extremely difficult for several reasons. For instance, a limited number of observed characters makes it difficult to generate unobserved characters with sufficient visual consistency. Also, different natural conditions, like brightness, contrast, shadow, perspective distortion, complex background, etc., make it harder to replace a character directly in an image. The main motivation of this work is to design an algorithm for editing textual information present in images in a convenient way similar to the conventional text editors.

Earlier, researchers have proposed font synthesis algorithms based on different geometrical features of fonts \cite{campbell2014learning,phan2015flexyfont,suveeranont2010example}. These geometrical models neither generalize the wide variety of available fonts nor can be applied directly to an image for character synthesis. Later, researchers have addressed the problem of generating unobserved characters of a particular font from some defined or random set of observations using deep learning algorithms \cite{baluja2017learning,chang2017chinese,upchurch2016z}. With the emergence of Generative Adversarial Network (GAN) models, the problem of character synthesis has also been addressed using GAN-based algorithms \cite{azadi2018multi,lyu2017auto}. Though GAN-based font synthesis could be used to estimate the target character, several challenges make the direct implementation of font synthesis for scene images difficult. Firstly, most of the GAN-based font synthesis models require an explicit recognition of the source character. As recognition of text in scene images is itself a challenging problem, it is preferable if the target characters can be generated without a recognition step. Otherwise, any error in the recognition process would accumulate, and make the entire text editing process unstable. Secondly, it is often observed that a particular word in an image may have a mixture of different font types, sizes, colors, etc. Even depending on the relative location of the camera and the texts in the scene, each character may experience a different amount of perspective distortion. Some GAN-based models \cite{azadi2018multi,lyu2017auto} require multiple observations of a font-type to faithfully generate unobserved characters. A multiple observation-based generation strategy requires a rigorous distortion removal step before applying generative algorithms. Thus, rather than a word-level generation, we follow a character-level generative model to accommodate maximum flexibility.

\textbf{Contributions}: To the best of our knowledge, this is the first work that attempts to modify texts in scene images. For this purpose, we design a generative network that adapts to the font features of a single character and generates other necessary characters. We also propose a model to transfer the color of the source character to the target character. The entire process works without any explicit character recognition. 

To restrict the complexity of our problem, we limit our discussion to the scene texts with upper-case non-overlapping characters. However, we demonstrate in Figs. \ref{fig:results} and \ref{fig:results_lowercase} that the proposed method can also be applied for lower-case characters and numerals.

\section{Related Works}
Because of its large potential, character synthesis from a few examples is a well-known problem. Previously, several pieces of work tried to address the problem using geometrical modeling of fonts \cite{campbell2014learning,phan2015flexyfont,suveeranont2010example}. Different synthesis models are also proposed by researchers explicitly for Chinese font generation \cite{lyu2017auto,zhou2011easy}. Along with statistical models \cite{phan2015flexyfont} and bilinear factorization \cite{tenenbaum2000separating}, machine learning algorithms are used to transfer font features. Recently, deep learning techniques also become popular in the font synthesis problem. Supervised \cite{upchurch2016z} and definite samples \cite{baluja2017learning} of observations are used to generate the unknown samples using deep neural architecture. Recently, Generative Adversarial Network (GAN) models are found to be effective in different image synthesis problems. GANs can be used in image style transfer \cite{gatys2016image}, structure generation \cite{isola2017image} or in both \cite{azadi2018multi}. Some of these algorithms achieved promising results in generating font structures \cite{chang2017chinese,lyu2017auto}, whereas some exhibits the potential to generate complex fonts with color \cite{azadi2018multi}. To the best of our knowledge, these generative algorithms work with text images that are produced using design software, and their applicability to edit real scene images are unknown. Moreover, most of the algorithms \cite{azadi2018multi,baluja2017learning} require explicit recognition of the source characters to generate the unseen character set. This may create difficulty in our problem as text recognition in scene images is itself a challenging problem \cite{bai2018edit,gupta2018learning,neumann2017scene} and any error in the recognition step may affect the entire generative process. Character generation from multiple observations is also challenging for scene images as the observed characters may have distinctively different characteristics like font types, sizes, colors, perspective distortions, etc.

Convolutional Neural Network (CNN) is proved to be effective in style transfer with generative models \cite{gatys2016image,li2016combining,liao2017visual}. Recently, CNN models are used to generate style and structure with different visual features \cite{dosovitskiy2016learning}. We propose a CNN-based character generation network that works without any explicit recognition of the source characters. For a natural-looking generation, it is also important to transfer the color and texture of the source character to the generated character. Color transfer is a widely explored topic in image processing \cite{reinhard2001color,tai2005local,welsh2002transferring}. Though these traditional approaches are good for transferring global colors in images, most of them are inappropriate for transferring colors in more localized character regions. Recently, GANs are also employed in color transfer problem \cite{azadi2018multi,li2018emerging}. In this work, we introduce a CNN-based color transfer model that takes the color information present in the source character and transfer it to the generated target character. The proposed color transfer model not only transfers solid colors from source to target character, it can also transfer gradient colors keeping subtle visual consistency.

\section{Methodology}
\begin{figure*}
\includegraphics[width=\textwidth]{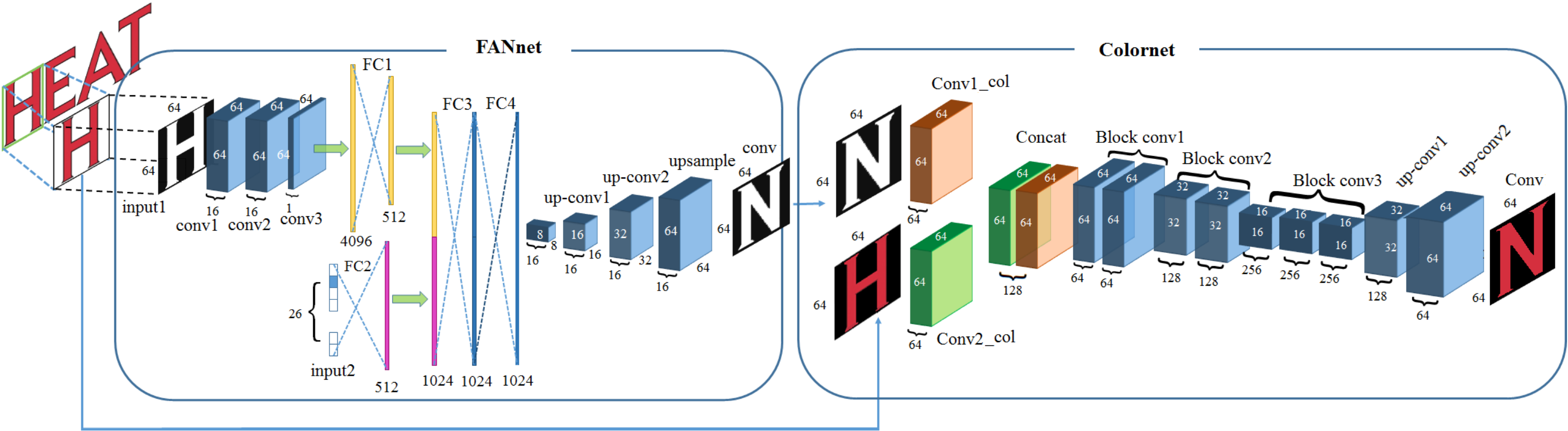}
\caption{Architecture of FANnet and Colornet. At first, the target character (`N') is generated from the source character (`H') by FANnet keeping structural consistency. Then, the source color is transferred to the target by Colornet preserving visual consistency. Layer names in the figure are: \textit{conv} = 2D convolution, \textit{FC} = fully-connected, \textit{up-conv} = upsampling + convolution.}
\label{fig:network_architecture}
\end{figure*}

\begin{figure}
\centering
\includegraphics[width=\linewidth]{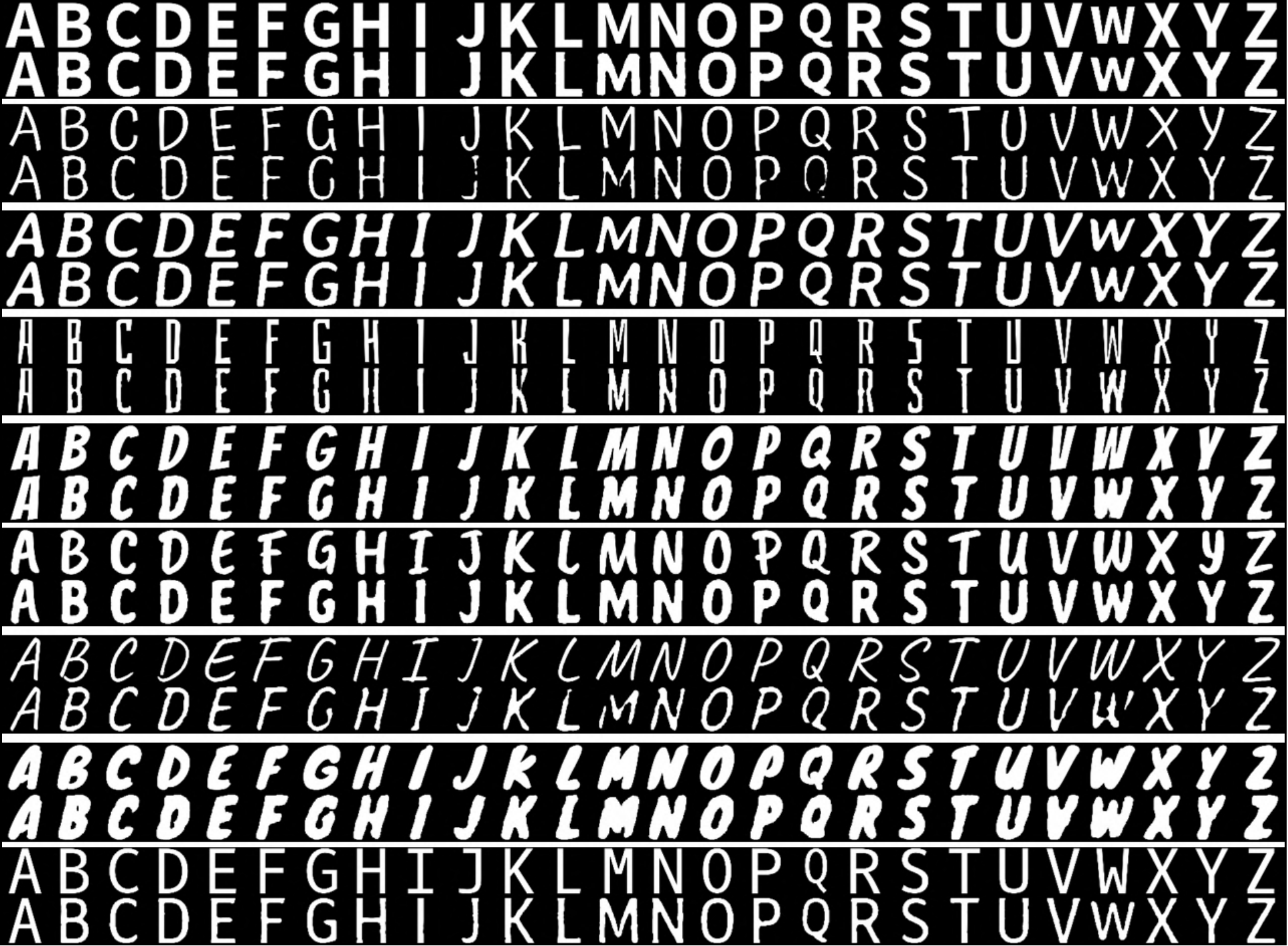}
\caption{Generation of target characters using FANnet. In each image block, the upper row shows the ground truth and the bottom row shows the generated characters when the network has observed one particular source character (`A') in each case.}
\label{fig:fannet_output}
\end{figure}

The proposed method is composed of the following steps: (1) Selection of the source character to be replaced, (2) Generation of the binary target character, (3) Color transfer and (4) Character placement. In the first step, we manually select the text area that requires to be modified. Then, the algorithm detects the bounding boxes of each character in the selected text region. Next, we manually select the bounding box around the character to be modified and also specify the target character. Based on these user inputs, the target character is generated, colorized and placed in the inpainted region of the source character.

\subsection{Selection of the source character}\label{sec:character_selection}
Let us assume that $I$ is an image that has multiple text regions, and $\Omega$ is the domain of a text region that requires modification. The region $\Omega$ can be selected using any text detection algorithm \cite{busta2017deep,lyu2018multi,yin2013robust}. Alternatively, a user can select the corner points of a polygon that bounds a word to define $\Omega$. In this work, we use EAST \cite{zhou2017east} to tentatively mark the text regions, followed by a manual quadrilateral corner selection to define $\Omega$. After selecting the text region, we apply the MSER algorithm \cite{chen2011robust} to detect the binary masks of individual characters present in the region $\Omega$. However, MSER alone cannot generate a sharp mask for most of the characters. Thus, we calculate the final binarized image $I_c$ defined as

\begin{equation*}
\begin{array}{r@{}l}
\begin{aligned}
I_c(\mathbf{p}) =
\begin{cases}
I_M(\mathbf{p})\bigodot I_B(\mathbf{p}) &\text{if } \mathbf{p}\in \Omega\\
0 & \text{otherwise}
\end{cases}
\end{aligned}
\end{array}
\end{equation*}
where $I_M$ is the binarized output of the MSER algorithm \cite{chen2011robust} when applied on $I$, $I_B$ is the binarized image of $I$ and $\bigodot$ denotes the element-wise product of matrices. The image $I_c$ contains the binarized characters in the selected region $\Omega$. If the color of the source character is darker than its background, we apply inverse binarization on $I$ to get $I_B$.

Assuming the characters are non-overlapping, we apply a connected component analysis and compute the minimum bounding rectangles of each connected component. If there are $N$ number of connected components present in a scene, $C_n\subseteq \Omega$ denotes the $n^{\text{th}}$ connected area where $0<n\leq N$. The bounding boxes $B_n$ contain the same indices as the connected areas that they are bounding. The user specifies the indices that they wish to edit. We define $\Theta$ as the set of indices that require modification, such that $|\Theta|\leq N$, where $|.|$ denotes the cardinality of a set. The binarized images $I_{C_\theta}$ associated with components $C_{\theta}$, $\theta\in \Theta$ are the source characters, and with proper padding followed by scaling (discussed in Sec. \ref{sec:character_generation}), they individually act as the input of the font generation network. Each $I_{C_\theta}$ has the same dimension with the bounding box $B_\theta$.

\begin{figure}
\centering
\includegraphics[width=\linewidth]{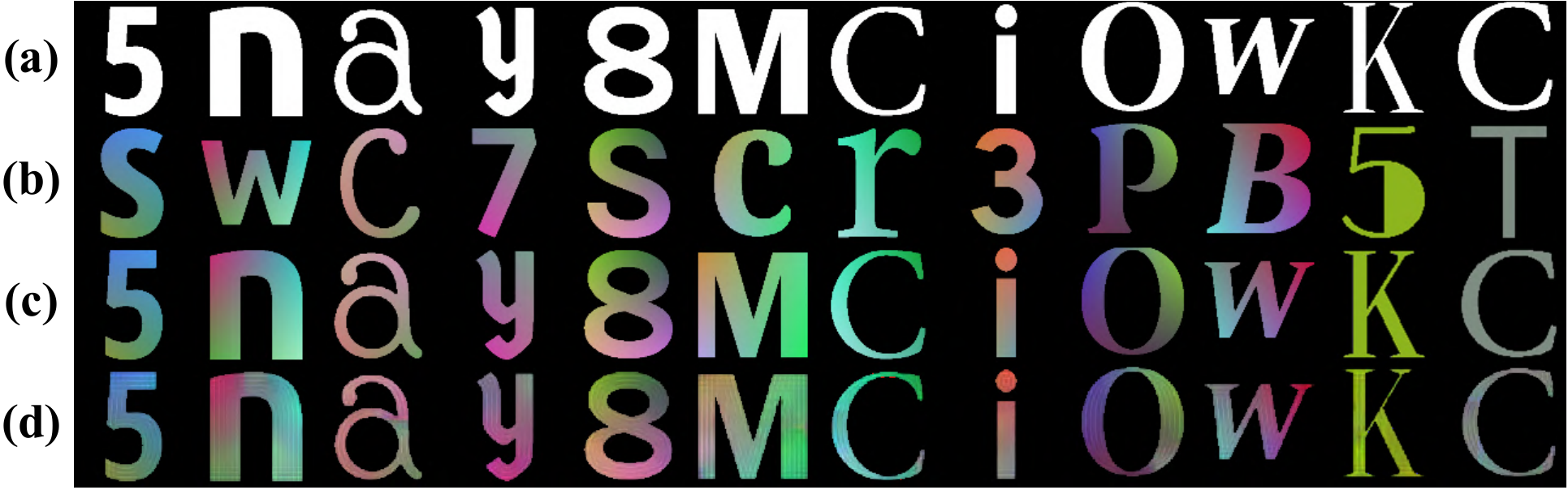}
\caption{Color transfer using Colornet: \textbf{(a)} Binary target character; \textbf{(b)} Color source character; \textbf{(c)} Ground truth; \textbf{(d)} Color transferred image. It can be observed that Colornet can successfully transfer solid color as well as gradient color.}
\label{fig:colornet_output}
\end{figure}

\subsection{Generation of the binary target character}\label{sec:character_generation}
Conventionally, most of the neural networks take square images as input. However, as $I_{C_\theta}$ may have different aspect ratios depending on the source character, font type, font size etc., a direct resizing of $I_{C_\theta}$ would distort the actual font features of the character. Rather, we pad $I_{C_\theta}$ maintaining its aspect ratio to generate a square binary image $I_\theta$ of size $m_\theta\times m_\theta$ such that, $m_\theta=\text{max}(h_\theta,w_\theta)$, where $h_\theta$ and $w_\theta$ are the height and width of bounding box $B_\theta$ respectively, and $\text{max}(.)$ is a mathematical operation that finds the maximum value. We pad both sides of $I_{C_\theta}$ along $x$ and $y$ axes with $p_x$ and $p_y$ respectively to generate $I_\theta$ such that
\begin{equation*}
p_x = \left[ {\frac{m_\theta-w_\theta}{2}} \right],\;\;
p_y = \left[ {\frac{m_\theta-h_\theta}{2}} \right]
\end{equation*}
followed by reshaping $I_\theta$ to a square dimension of $64\times 64$.

\subsubsection{Font Adaptive Neural Network (FANnet)}\label{sec:fannet}
Our generative font adaptive neural network (FANnet) takes two different inputs -- an image of the source character of size $64\times 64$ and a one-hot encoding $\mathbf{v}$ of length 26 of the target character. For example, if our target character is `H', then $\mathbf{v}$ has the value 1 at index 7 and 0 in every other location. The input image passes through three convolution layers having 16, 16 and 1 filters respectively, followed by flattening and a fully-connected (FC) layer FC1. The encoded vector $\mathbf{v}$ also passes through an FC layer FC2. The outputs of FC1 and FC2 give 512 dimensional latent representations of respective inputs. Outputs of FC1 and FC2 are concatenated and followed by two more FC layers, FC3 and FC4 having 1024 neurons each. The expanding part of the network contains reshaping to a dimension $8\times 8 \times 16$ followed by three `up-conv' layers having 16, 16 and 1 filters respectively. Each `up-conv' layer contains an upsampling followed by a 2D convolution. All the convolution layers have kernel size $3\times 3$ and ReLU activation. The architecture of FANnet is shown in Fig. \ref{fig:network_architecture}. The network minimizes the mean absolute error (MAE) while training with Adam optimizer \cite{kingma2015adam} with learning rate $lr=10^{-3}$, momentum parameters $\beta_1=0.9$, $\beta_2=0.999$ and regularization parameter $\epsilon=10^{-7}$.

We train FANnet with 1000 fonts with all 26 upper-case character images as inputs and 26 different one-hot encoded vectors for each input. It implies that for 1000 fonts, we train the model to generate any of the 26 upper-case target character images from any of the 26 upper-case source character images. Thus, our training dataset has a total of 0.6760 million input pairs. The validation set contains 0.2028 million input pairs generated from another 300 fonts. We select all the fonts from Google Fonts database \cite{googlefonts}. We apply the Otsu thresholding technique \cite{otsu1979threshold} on the grayscale output image of FANnet to get a binary target image.

\subsection{Color transfer}\label{sec:colornet}
It is important to have a faithful transfer of color from the source character for a visually consistent generation of the target character. We propose a CNN based architecture, named Colornet, that takes two images as input -- colored source character image and binary target character image. It generates the target character image with transferred color from the source character image. Each input image goes through a 2D convolution layer, Conv1\_col and Conv2\_col for input1 and input2 respectively. The outputs of Conv1\_col and Conv2\_col are batch-normalized and concatenated, which is followed by three blocks of convolution layers with two max-pooling layers in between. The expanding part of Colornet contains two `up-conv' layers followed by a 2D convolution. All the convolution layers have kernel size $3\times 3$ and Leaky-ReLU activation with $\alpha=0.2$. The architecture of Colornet is shown in Fig. \ref{fig:network_architecture}. The network minimizes the mean absolute error (MAE) while training with Adam optimizer that has the same parameter settings as mentioned in Sec. \ref{sec:fannet}.

We train Colornet with synthetically generated image pairs. For each image pair, the color source image and the binary target image both are generated using the same font type randomly selected from 1300 fonts. The source color images contain both solid and gradient colors so that the network can learn to transfer a wide range of color variations. We perform a bitwise-AND between the output of Colornet and the binary target image to get the final colorized target character image.

\subsection{Character placement}\label{sec:character_placement}
Even after the generation of target characters, the placement requires several careful operations. First, we need to remove the source character from $I$ so that the generated target character can be placed. We use image inpainting \cite{telea2004image} using $W(I_{C_\theta},\psi)$ as a mask to remove the source character, where $W(I_b,\psi)$ is the dilation operation on any binary image $I_b$ using the structural element $\psi$. In our experiments, we consider $\psi=3\times 3$. To begin the target character placement, first the output of Colornet is resized to the dimension of $I_\theta$. We consider that the resized color target character is $R_\theta$ with minimum rectangular bounding box $B^R_\theta$. If $B^R_\theta$ is smaller or larger than $B_\theta$, then we need to remove or add the region $B_\theta \setminus B^R_\theta$ accordingly so that we have the space to position $R_\theta$ with proper inter-character spacing. We apply the content-aware seam carving technique \cite{avidan2007seam} to manipulate the non-overlapping region. It is important to mention that if $B^R_\theta$ is smaller than $B_\theta$ then after seam carving, the entire text region $\Omega$ will shrink to a region $\Omega_s$, and we also need to inpaint the region $\Omega \setminus \Omega_s$ for consistency. However, both the regions $B_\theta \setminus B^R_\theta$ and $\Omega \setminus \Omega_s$ are considerably small and are easy to inpaint for upper-case characters. Finally, we place the generated target character on the seam carved image such that the centroid of $B^R_\theta$ overlaps with the centroid of $B_\theta$.

\section{Results}
\begin{figure*}
\includegraphics[width=\textwidth]{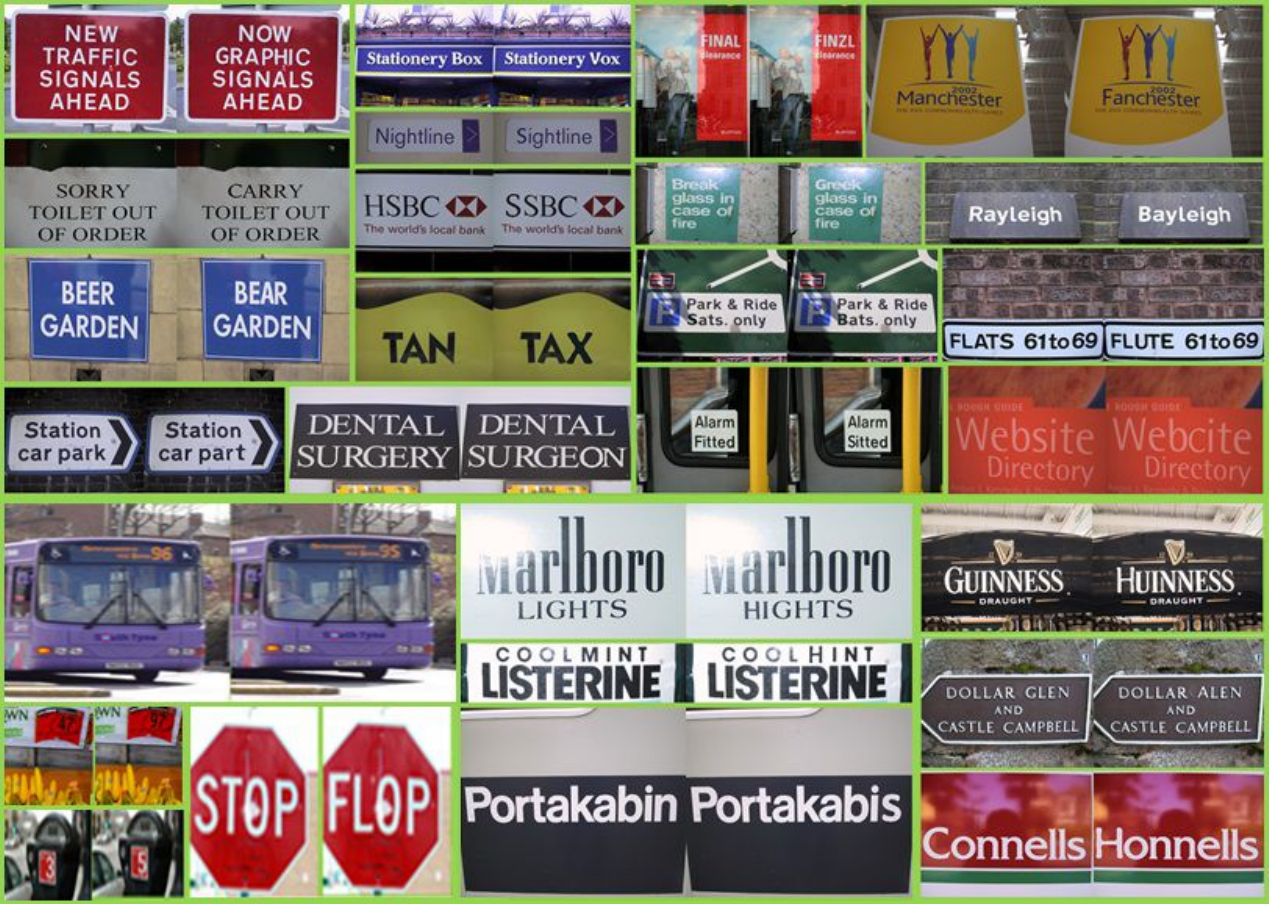}
\caption{Images edited using STEFANN. In each image pair, the left image is the original image and the right image is the edited image. It can be observed that STEFANN can faithfully edit texts even in the presence of specular reflection, shadow, perspective distortion, etc. It is also possible to edit lower-case characters and numerals in a scene image. STEFANN can easily edit multiple characters and multiple text regions in an image. \textbf{More results are included in the supplementary materials.}}
\label{fig:results}
\end{figure*}

We tested\footnote{Code: \url{https://github.com/prasunroy/stefann}} our algorithm on COCO-Text and ICDAR datasets. The images in the datasets are scene images with texts written with different unknown fonts. In Fig. \ref{fig:results}, we show some of the images that are edited using STEFANN. In each image pair, the left image is the original image and the right image is the edited image. In some of the images, several characters are edited in a particular text region, whereas in some images, several text regions are edited in a single image. It can be observed that not only the font features and colors are transferred successfully to the target characters, but also the inter-character spacing is maintained in most of the cases. Though all the images are natural scene images and contain different lighting conditions, fonts, perspective distortions, backgrounds, etc., in all the cases STEFANN is able to edit the images without any significant visual inconsistency.

\begin{figure}
\centering
\includegraphics[width=\linewidth]{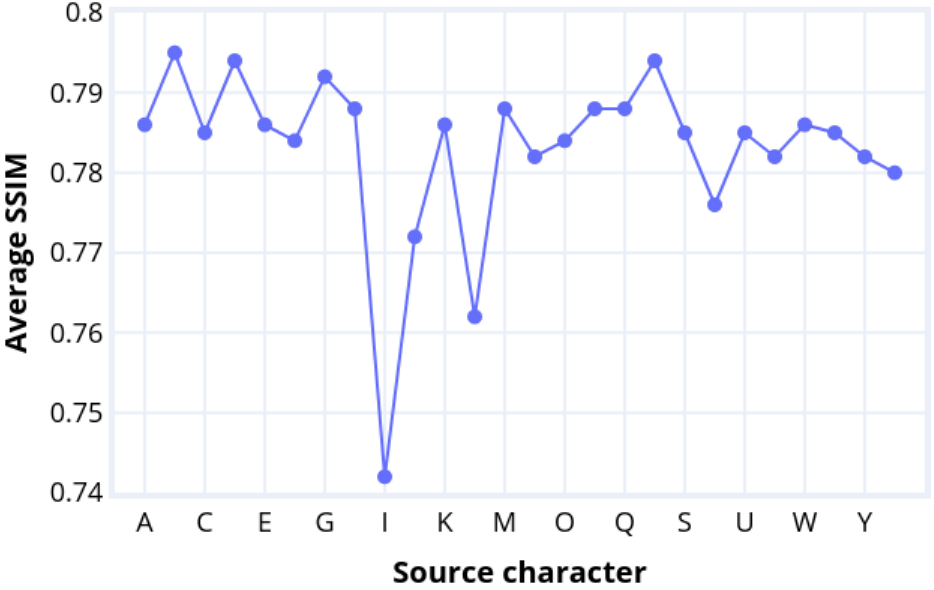}
\caption{Average SSIM of the generated characters for each different source character.}
\label{fig:fannet_ssim}
\end{figure}

\textbf{Evaluation and ablation study of FANnet:}
To evaluate the performance of the proposed FANnet model, we take one particular source character and generate all possible target characters. We repeat this process for every font in the test set. The outputs for some randomly selected fonts are shown in Fig. \ref{fig:fannet_output}. Here, we only provide an image of character `A' as the source in each case and generate all 26 characters. To quantify the generation quality of FANnet, we select one source character at a time as input and measure the average structural similarity index (ASSIM) \cite{wang2004image} of all 26 generated target characters against respective ground truth images over a set of 300 test fonts. In Fig. \ref{fig:fannet_ssim}, we show the average SSIM of the generated characters for each different source character. It can be seen from the ASSIM scores that some characters, like `I' and `L', are less informative as they produce lower ASSIM values, whereas characters, like `B' and `M', are structurally more informative in the generation process.

\begin{table}
\caption{Ablation study of FANnet architecture. Layer names are similar to Fig. \ref{fig:network_architecture}.}
\centering
\begin{tabular}{|l|l|l|}
\hline
\textbf{\#} & \textbf{Excluded layer(s)} & \textbf{ASSIM}  \\ \hline
1           & up-conv1 + up-conv2        & 0.5664          \\ \hline
2           & FC4 + up-conv2             & 0.6426          \\ \hline
3           & FC1 + FC2 + up-conv2       & 0.6718          \\ \hline
4           & None (Proposed FANnet)     & \textbf{0.7712} \\ \hline \hline
5           & All (FCN)                  & 0.1332          \\ \hline
\end{tabular}
\label{tab:fannet_ablation}
\end{table}

\begin{figure}
\centering
\includegraphics[width=\linewidth]{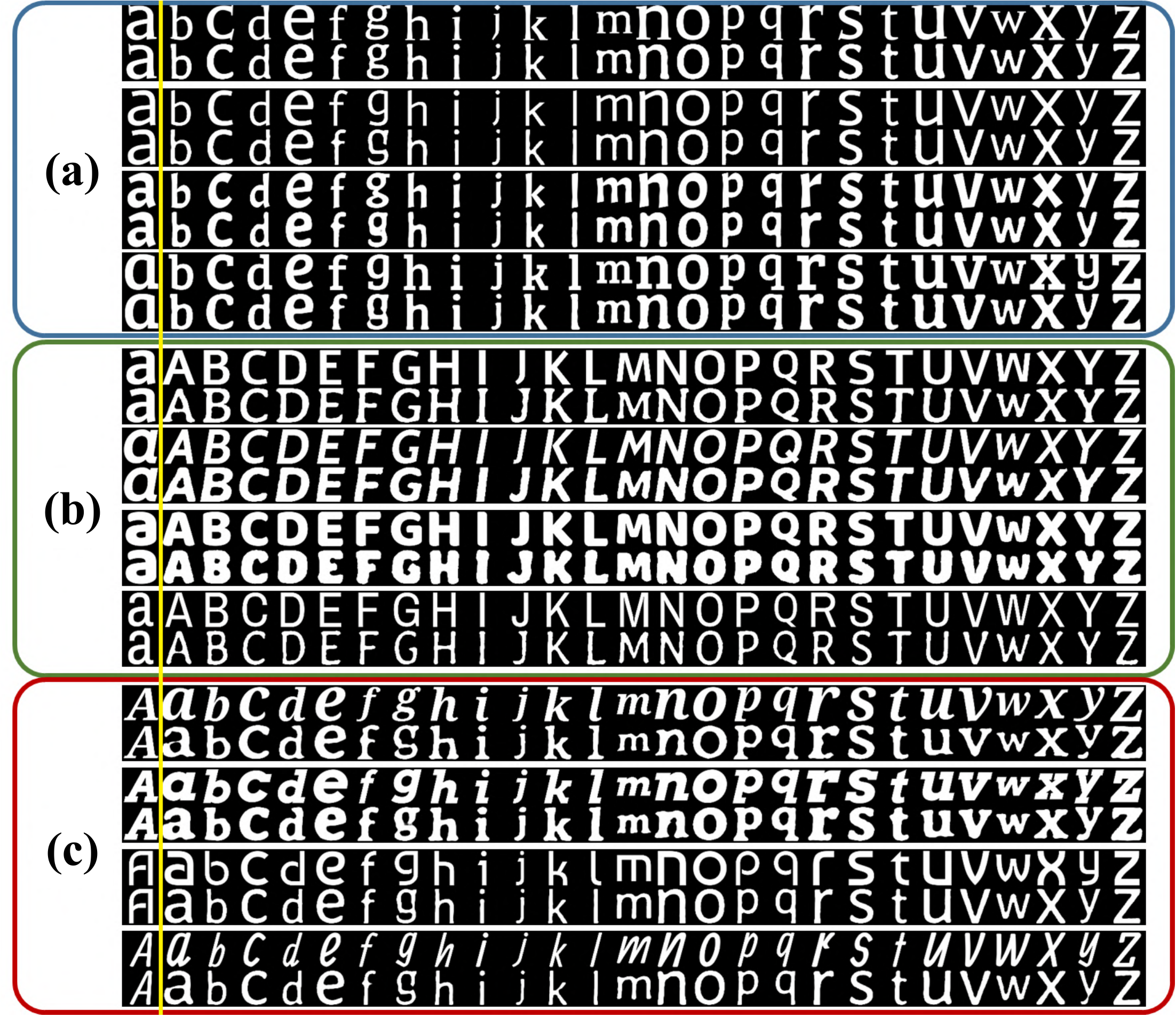}
\caption{Additional generation results of target characters using FANnet: In each image block, the upper row shows the ground truth, and the bottom row shows the generated characters when the network observes only one particular character: \textbf{(a)} lower-case target characters generated from a lower-case source character (`a'); \textbf{(b)} upper-case target characters generated from a lower-case source character (`a'); \textbf{(c)} lower-case target characters generated from an upper-case source character (`A').}
\label{fig:fannet_generation}
\end{figure}

We perform ablation studies using ASSIM score to validate the proposed FANnet architecture. The results are shown in Table \ref{tab:fannet_ablation}. As Fully Convolutional Networks (FCN) are often used in generative models, we tried to develop FANnet using FCN models at first. Unfortunately, none of the developed FCN architecture for FANnet worked properly. To demonstrate it, we have included an FCN architecture in Table \ref{tab:fannet_ablation} which is similar to ColorNet with two inputs -- one is the source character image and another is the target character in standard `Times New Roman' font. Table \ref{tab:fannet_ablation} clearly shows the motivation of the proposed FANnet model as it achieves the highest ASSIM score in the ablation study. In our ablation study, when the `up-conv' layer is removed, it is replaced with an upsampling layer by a factor of 2 to maintain the image size. During the ablation study, replacement of ReLU activation with Leaky-ReLU activation gives an ASSIM score of 0.4819. To further analyze the robustness of the generative model, we build another network with the same architecture as described in Sec. \ref{sec:fannet}, and train it with lower-case character images of the same font database. The output of the model is shown in Fig. \ref{fig:fannet_generation}(a) when only a lower-case character (character `a') is given as the source image and all the lower-case characters are generated. As shown in Fig. \ref{fig:fannet_generation}(b) and Fig. \ref{fig:fannet_generation}(c), we also observe that the model can transfer some font features even when a lower-case character is provided as the input and we try to generate the upper-case characters or vice versa.

\begin{figure}
\centering
\includegraphics[width=\linewidth]{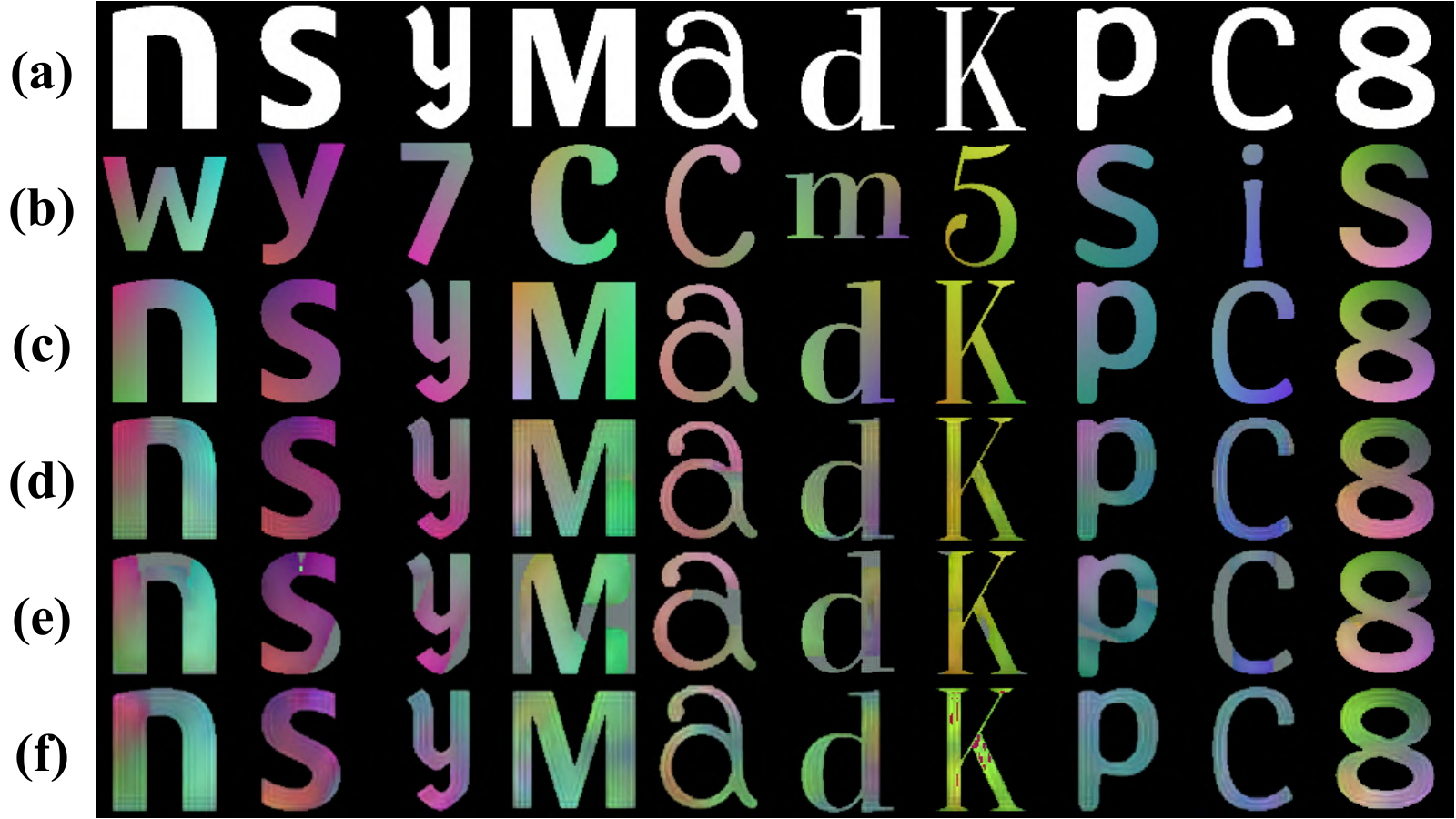}
\caption{Color transfer results for different models: \textbf{(a)} Binary target character; \textbf{(b)} Color source character; \textbf{(c)} Ground truth; \textbf{(d)} Output of the proposed Colornet model; \textbf{(e)} Output of Colornet-L; \textbf{(f)} Output of Colornet-F. It can be observed that the Colornet architecture discussed in Sec. \ref{sec:colornet} transfers the color from source to target without any significant visual distortion.}
\label{fig:colornet_generation}
\end{figure}

\textbf{Evaluation and ablation study of Colornet:}
The performance of the proposed Colornet is shown in Fig. \ref{fig:colornet_output} for both solid and gradient colors. It can be observed that in both cases, Colornet can faithfully transfer the font color of source characters to target characters. As shown in Fig. \ref{fig:colornet_output}, the model works equally well for all alphanumeric characters including lower-case and upper-case letters. To understand the functionality of the layers in Colornet, we perform an ablation study and select the best model architecture that faithfully transfers the color. We compare the proposed Colornet architecture with two other variants -- Colornet-L and Colornet-F. In Colornet-L, we remove `Block conv3' and `up-conv1' layers to perform layer ablation. In Colornet-F, we reduce the number of convolution filters to 16 in both `Conv1\_col' and `Conv2\_col' layers to perform filter ablation. The results of color transfer for all three Colornet variants are shown in Fig. \ref{fig:colornet_generation}. It can be observed that Colornet-L produces visible color distortion in the generated images, whereas some color information are not present in the images generated by Colornet-F.

\begin{figure*}
\centering
\includegraphics[width=\linewidth]{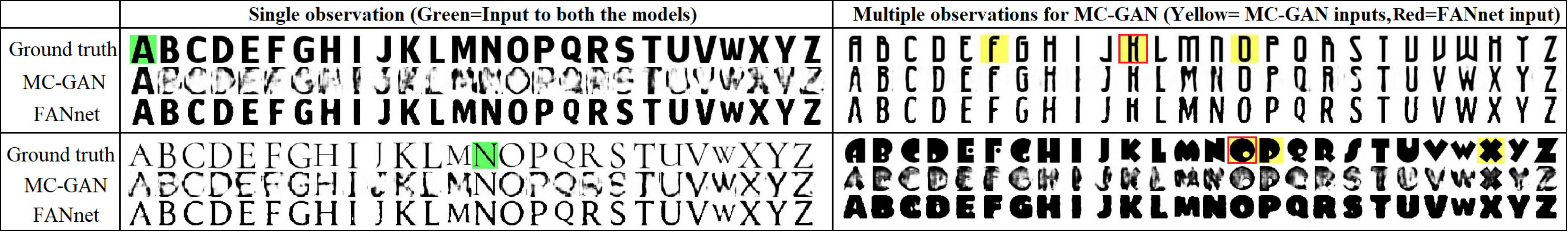}
\caption{Comparison between MC-GAN and the proposed FANnet architecture. The green color indicates input to both the models when only one observation is available. Yellow colors indicate input to MC-GAN and the red box indicates input to FANnet when 3 random observations are available. The evaluation is performed on the MC-GAN dataset.}
\label{fig:comparison_mcgan}
\end{figure*}

\begin{table}
\caption{Comparison of synthetic character generation between MC-GAN and FANnet.}
\scriptsize{
\begin{center}{
\begin{tabular}{|l|l|l|c|l|l|}
\hline
\multicolumn{2}{|c|}{\begin{tabular}[c]{@{}c@{}}MC-GAN\\ (1 observation)\end{tabular}} & \multicolumn{2}{c|}{\begin{tabular}[c]{@{}c@{}}MC-GAN\\ (3 random observations)\end{tabular}} & \multicolumn{2}{c|}{\begin{tabular}[c]{@{}c@{}}FANnet\\ (1 observation)\end{tabular}} \\ \hline
nRMSE & ASSIM & \multicolumn{1}{c|}{nRMSE} & \multicolumn{1}{c|}{ASSIM} & nRMSE & ASSIM \\ \hline
0.4568 & 0.4098 & 0.3628 & 0.5485 & 0.4504 & 0.4614 \\ \hline
\end{tabular}}
\end{center}}
\label{tab:comparison_mcgan}
\end{table}

\textbf{Comparison with other methods:}
To the best of our knowledge, there is no significant prior work that aims to edit textual information in natural scene images directly. MC-GAN \cite{azadi2018multi} is a recent font synthesis algorithm, but to apply it on scene text a robust recognition algorithm is necessary. Thus on many occasions, it is not possible to apply and evaluate its performance on scene images. However, the generative performance of the proposed FANnet is compared with MC-GAN as shown in Fig. \ref{fig:comparison_mcgan}. We observe that given a single observation of a source character, FANnet outperforms MC-GAN, but as the number of observations increases, MC-GAN performs better than FANnet. This is also shown in Table \ref{tab:comparison_mcgan} where we measure the quality of the generated characters using nRMSE and ASSIM scores. The comparison is done on the dataset \cite{azadi2018multi} which is originally used to train MC-GAN with multiple observations. But in this case, FANnet is not re-trained on this dataset which also shows the adaptive capability of FANnet. In another experiment, we randomly select 1000 fonts for training and 300 fonts for testing from the MC-GAN dataset. When we re-train FANnet with this new dataset, we get an ASSIM score of 0.4836 over the test set. For MC-GAN, we get ASSIM scores of 0.3912 (single observation) and 0.5679 (3 random observations) over the same test set.

\begin{figure}
\centering
\includegraphics[width=\linewidth]{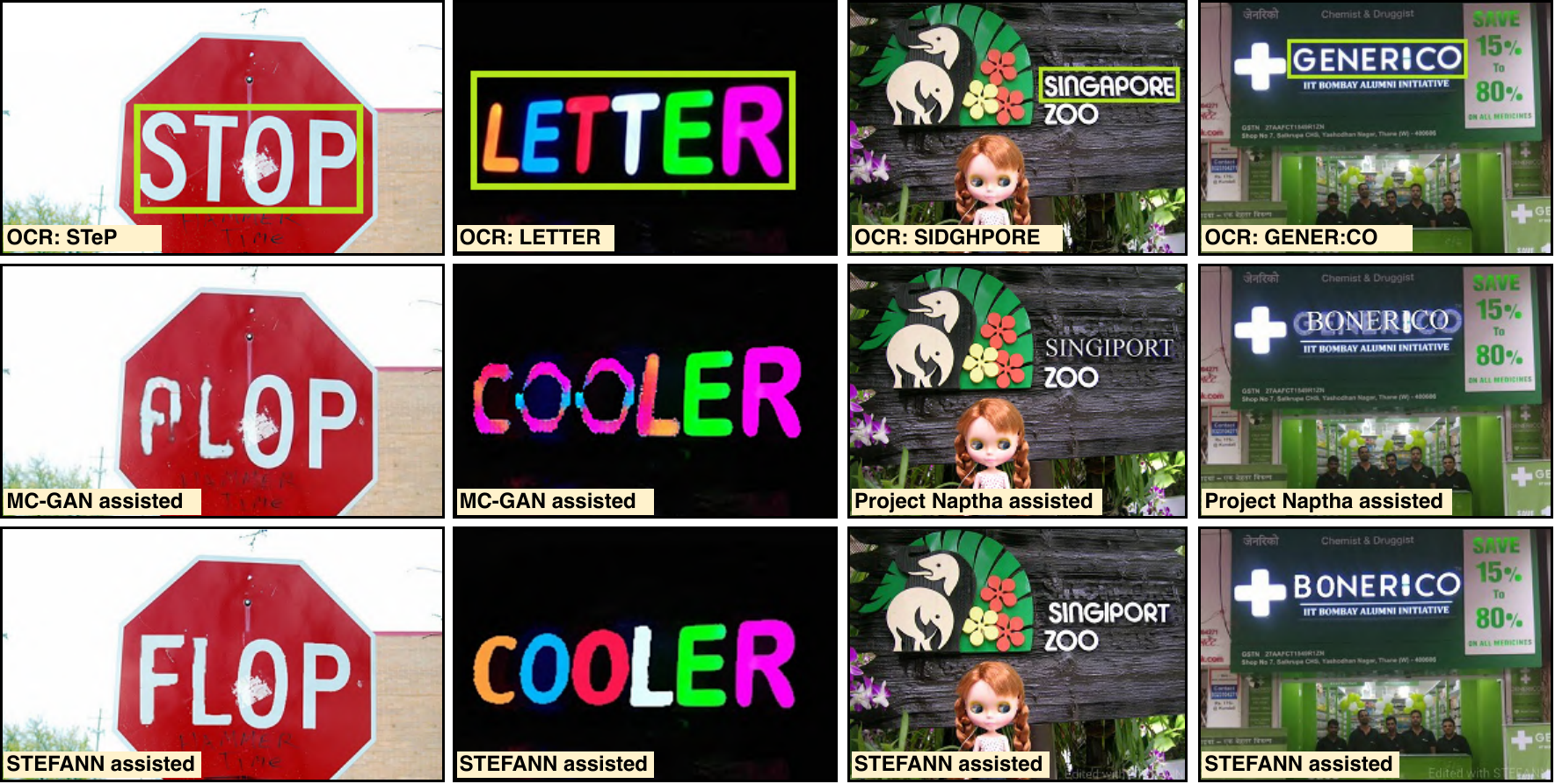}
\caption{Comparison among MC-GAN, Project Naptha and STEFANN assisted text editing on scene images. \textbf{Top row:} Original images. Text regions to be edited are highlighted with green bounding boxes. OCR predictions for these text regions are shown in respective insets. \textbf{Middle row:} MC-GAN or Project Naptha assisted text editing. \textbf{Bottom row:} STEFANN assisted text editing.}
\label{fig:comparison_others}
\end{figure}

We also perform a comparison among MC-GAN \cite{azadi2018multi}, Project Naptha \cite{projectnaptha} and STEFANN assisted text editing schemes on scene images as shown in Fig. \ref{fig:comparison_others}. For MC-GAN assisted editor, we replace FANnet and Colornet with MC-GAN cascaded with Tesseract v4 OCR engine \cite{smith2007overview}. Project Naptha provides a web browser extension that allows users to manipulate texts in images. It can be observed that the generative capability of MC-GAN is directly affected by recognition accuracy of the OCR and variation of scale and color among source characters, whereas Project Naptha suffers from weak font adaptability and inpainting.

To understand the perceptual quality of the generated characters, we take opinions from 115 users for 50 different fancy fonts randomly taken from the MC-GAN dataset to evaluate the generation quality of MC-GAN and FANnet. For a single source character, 100\% of users opine that the generation of FANnet is better than MC-GAN. For 3 random source characters, 67.5\% of users suggest that the generation of FANnet is preferable over MC-GAN.

\begin{figure}
\centering
\includegraphics[width=\linewidth]{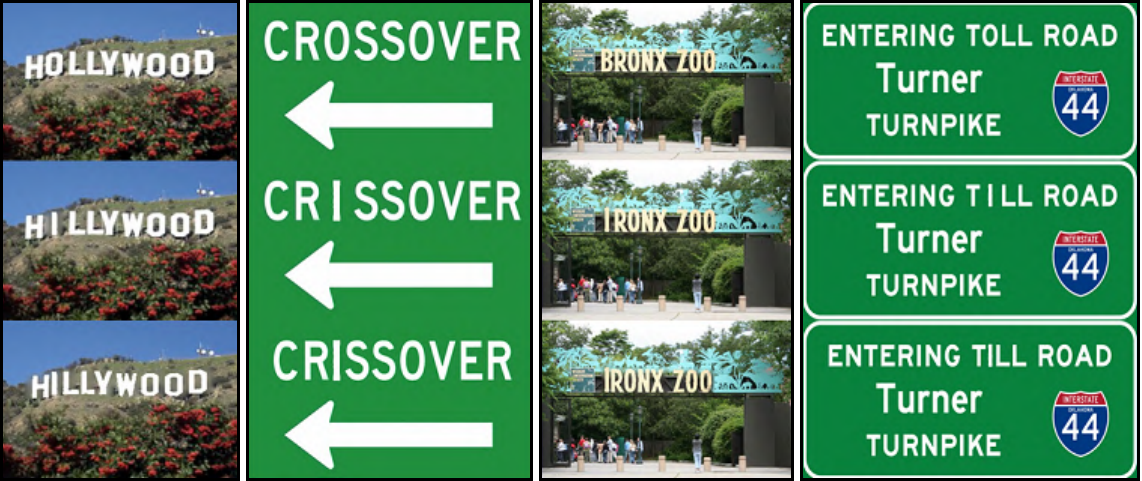}
\caption{Effectiveness of seam carving. In each column, top row shows the original image, middle row shows the edited image without seam carving and bottom row shows the edited image with seam carving.}
\label{fig:seam_carving}
\end{figure}

\textbf{Visual consistency during character placement:}
Automatic seam carving of the text region is an important step to perform perceptually consistent modifications maintaining the inter-character distance. Seam carving is particularly required when the target character is `I'. It can be seen from Fig. \ref{fig:seam_carving} that the edited images with seam carving look visually more consistent than those without seam carving.

\textbf{Evaluation of overall generation quality:}
To evaluate the overall quality of editing, we perform two opinion based evaluations with 136 viewers. First, they are asked to detect whether the displayed image is edited or not from a set of 50 unedited and 50 edited images shown at random. We get 15.6\% true-positive (TP), 37.1\% true-negative (TN), 12.8\% false-positive (FP) and 34.4\% false-negative (FN), which shows that the response is almost random. Next, the same viewers are asked to mark the edited character(s) from another set of 50 edited images. In this case, only 11.6\% of edited characters are correctly identified by the viewers.

\section{Discussion and Conclusion}
\begin{figure}
\centering
\includegraphics[width=\linewidth]{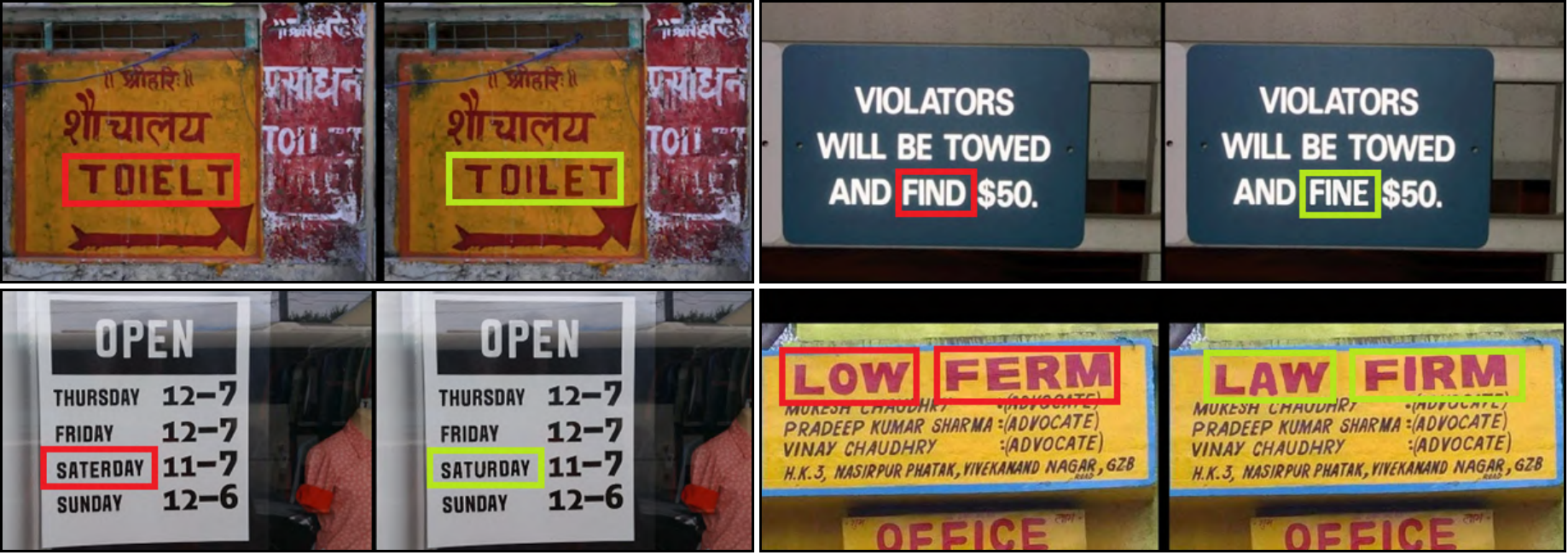}
\caption{Application of STEFANN. Misspelled words (bounded in Red) are corrected (bounded in Green) in scene images.}
\label{fig:results_corrected}
\end{figure}

\begin{figure}
\centering
\includegraphics[width=\linewidth]{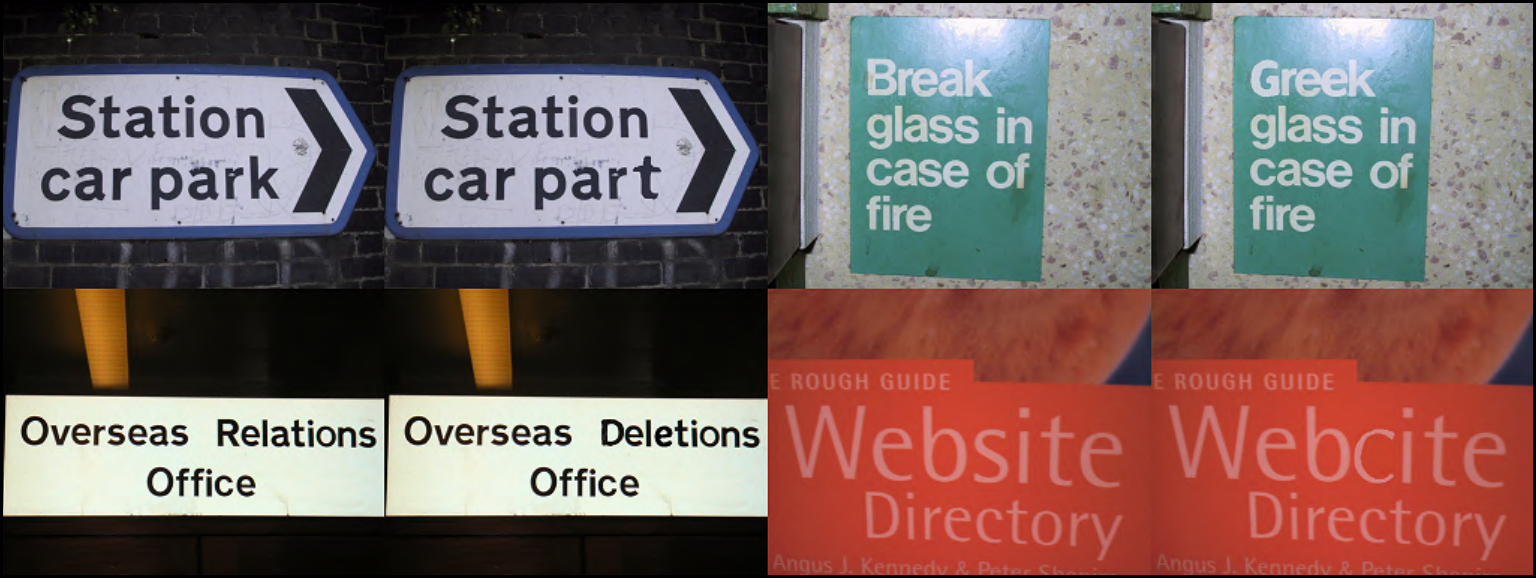}
\caption{Some images where lower-case characters are edited using STEFANN.}
\label{fig:results_lowercase}
\end{figure}

\begin{figure}
\centering
\includegraphics[width=\linewidth]{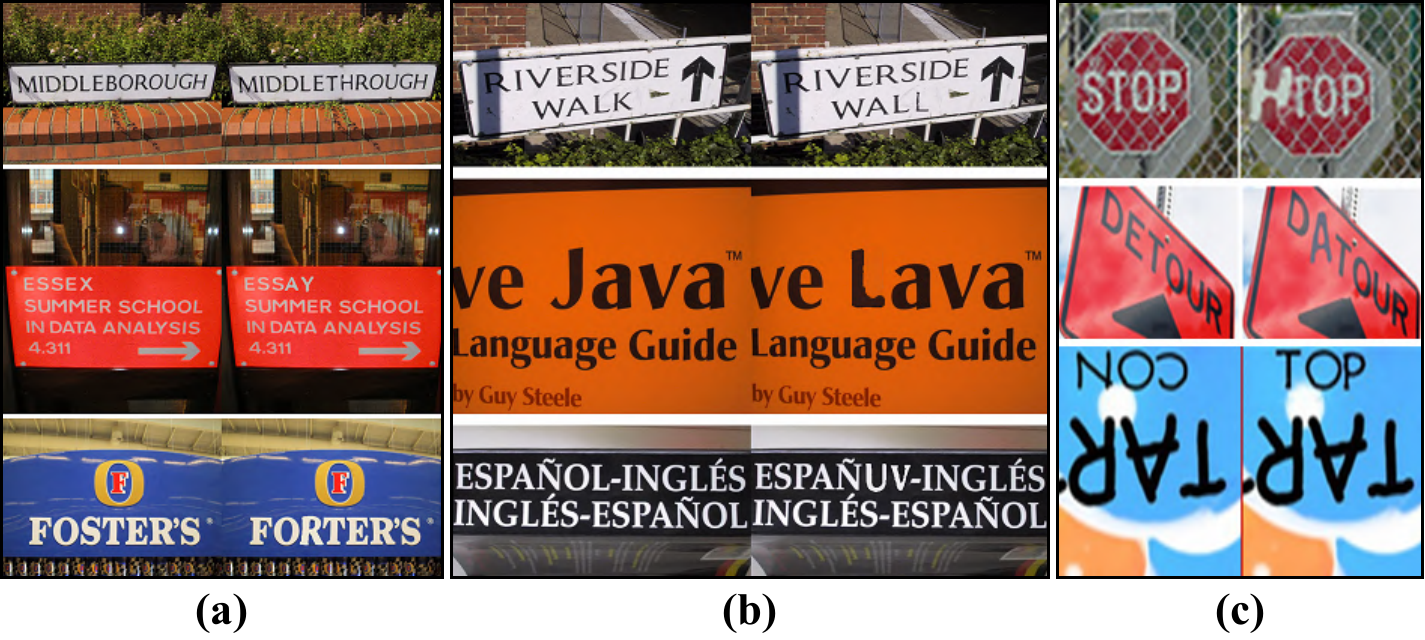}
\caption{Some images where STEFANN fails to edit text with sufficient visual consistency.}
\label{fig:results_failed}
\end{figure}

The major objective of STEFANN is to perform image editing for error correction, text restoration, image reusability, etc. Few such use cases are shown in Fig. \ref{fig:results_corrected}. Apart from these, with proper training, it can be used in font adaptive image-based machine translation and font synthesis. To the best of our knowledge, this is the first attempt to develop a unified platform to edit texts directly in images with minimal manual effort. STEFANN handles scene text editing efficiently for single or multiple characters while preserving visual consistency and maintaining inter-character spacing. Its performance is affected by extreme perspective distortion, high occlusion, large rotation, etc. which is expected as these effects are not present in the training data. Also, it should be noted that while training FANnet, we use Google Fonts \cite{googlefonts} which contains a limited number of artistic fonts, and we train Colornet with only solid and gradient colors. Thus at present, STEFANN does not accommodate editing complex artistic fonts or unusual texture patterns. The proposed FANnet architecture can also be used to generate lower-case target characters with similar architecture as discussed in Sec. \ref{sec:fannet}. However in the case of lower-case characters, it is difficult to predict the size of the target character only from the size of the source character. It is mainly because lower-case characters are placed in different `text zones' \cite{pal1999automatic} and the source character may not be replaced directly if the target character falls into a different text zone. In Fig. \ref{fig:results_lowercase}, we show some images where we edit the lowercase characters with STEFANN. In Fig. \ref{fig:results_failed}, we also show some cases where STEFANN fails to edit the text faithfully. The major reason behind the failed cases is an inappropriate generation of the target character. In some cases, the generated characters are not consistent with the same characters present in the scene [Fig. \ref{fig:results_failed}(a)], whereas in some cases the font features are not transferred properly [Fig. \ref{fig:results_failed}(b)]. We also demonstrate that STEFANN currently fails to work with extreme perspective distortion, high occlusion or large rotation [Fig. \ref{fig:results_failed}(c)]. In all the editing examples shown in this paper, the number of characters in a text region is not changed. One of the main limitations of the present methodology is that the font generative model FANnet generates images with dimension $64\times 64$. While editing high-resolution text regions, a rigorous upsampling is often required to match the size of the source character. This may introduce severe distortion of the upsampled target character image due to interpolation. In the future, we plan to integrate super-resolution \cite{wang2018recovering,wang2018esrgan} to generate very high-resolution character images that are necessary to edit any design or illustration. Also, we use MSER to extract text regions for further processing. So, if MSER fails to extract the character properly, the generation results will be poor. However, this can be rectified using better character segmentation algorithms. It is worth mentioning that robust image authentication and digital forensic techniques should be integrated with such software to minimize the risk of probable misuses of realistic text editing in images.

\section*{Acknowledgements}
We would like to thank NVIDIA Corporation for providing a TITAN X GPU through the GPU Grant Program.

%%%%%%%%%% REFERENCES
{\small
\bibliographystyle{ieee_fullname}
\bibliography{references}
}

%%%%%%%%%% APPENDIX
\clearpage
% \documentclass[10pt,letterpaper]{article}

% \usepackage{cvpr}
% \usepackage{times}
% \usepackage{epsfig}
% \usepackage{graphicx}
% \usepackage{amsmath}
% \usepackage{amssymb}
% \usepackage[pagebackref=true,colorlinks=true,breaklinks=true,bookmarks=false,citecolor=blue]{hyperref}

% \cvprfinalcopy

% \def\cvprPaperID{8915}
% \def\httilde{\mbox{\tt\raisebox{-.5ex}{\symbol{126}}}}

% \ifcvprfinal\pagestyle{empty}\fi
% % \setcounter{page}{4321}

% \begin{document}

% %%%%%%%%%% TITLE
% \title{Supplementary Materials\\for\\STEFANN: Scene Text Editor using Font Adaptive Neural Network}
% \author{
%   Prasun Roy$^{1}$\thanks{These authors contributed equally to this work.}~,~~
%   Saumik Bhattacharya$^{2}$\footnotemark[1]~,~~
%   Subhankar Ghosh$^{1}$\footnotemark[1]~,~~
%   and Umapada Pal$^{1}$ \\
%   $^{1}$Indian Statistical Institute, Kolkata, India \\
%   $^{2}$Indian Institute of Technology, Kharagpur, India \\
%   % {\tt\small \{prasunroy.pr, soumiksweb, sgcs2005\}@gmail.com, umapada@isical.ac.in} \\
%   {\tt\small \url{https://prasunroy.github.io/stefann}} \\
% }
% \maketitle
% \thispagestyle{empty}

%%%%%%%%%% BODY
\renewcommand{\thefigure}{S\arabic{figure}}
\setcounter{figure}{0}

\onecolumn
\section*{\centering \LARGE \textbf{Supplementary Materials}}

\vspace{2em}

\begin{figure}[h]
\centering
\includegraphics[width=\textwidth]{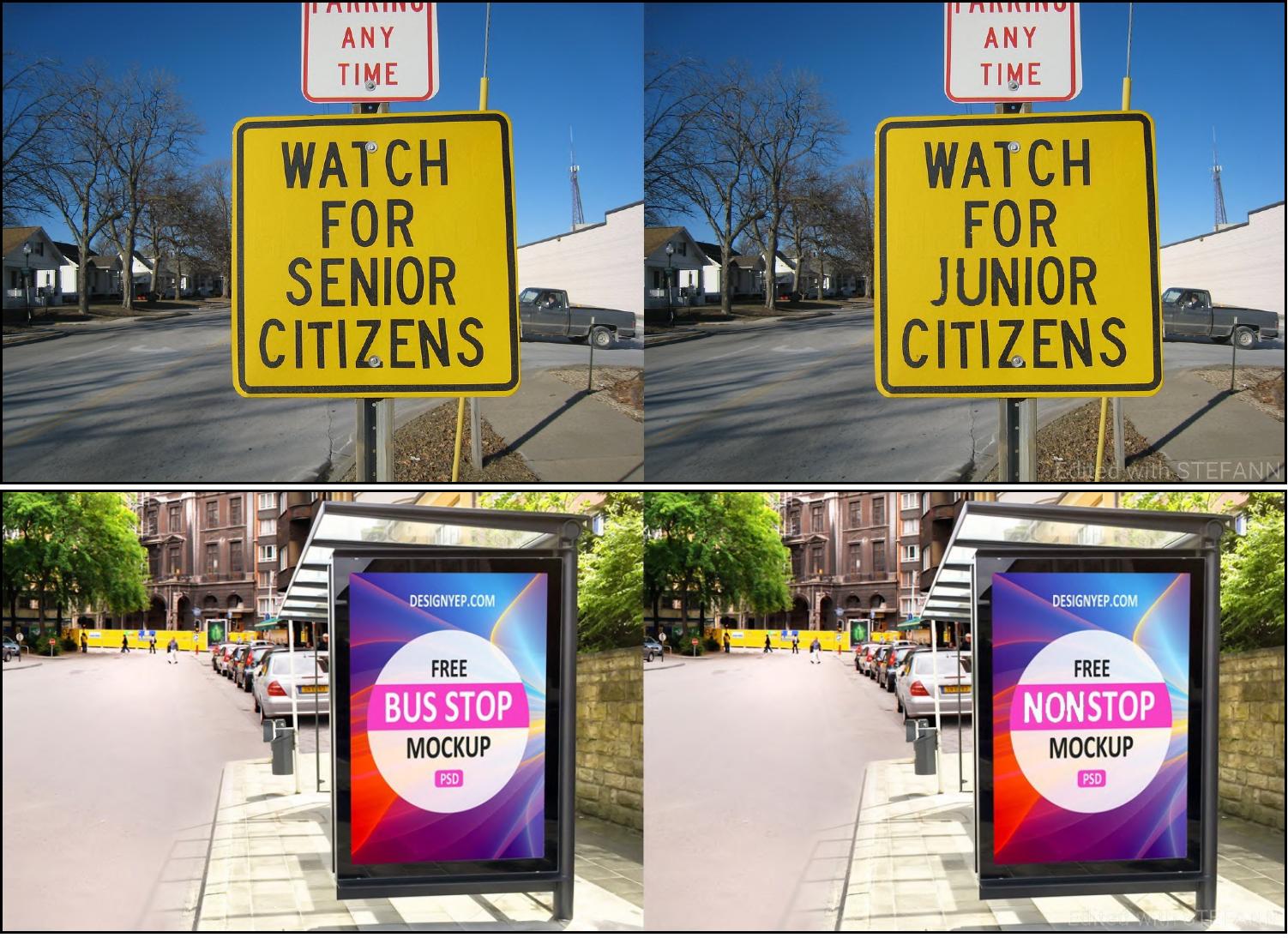}
\caption{Editing texts in road signboards with STEFANN. \textbf{Left:} Original images. \textbf{Right:} Edited images. Text regions are intentionally left unmarked to show the visual coherence of edited texts with original texts without creating passive attention.}
\label{fig:fig1}
\end{figure}

\begin{figure*}
\centering
\includegraphics[width=\textwidth]{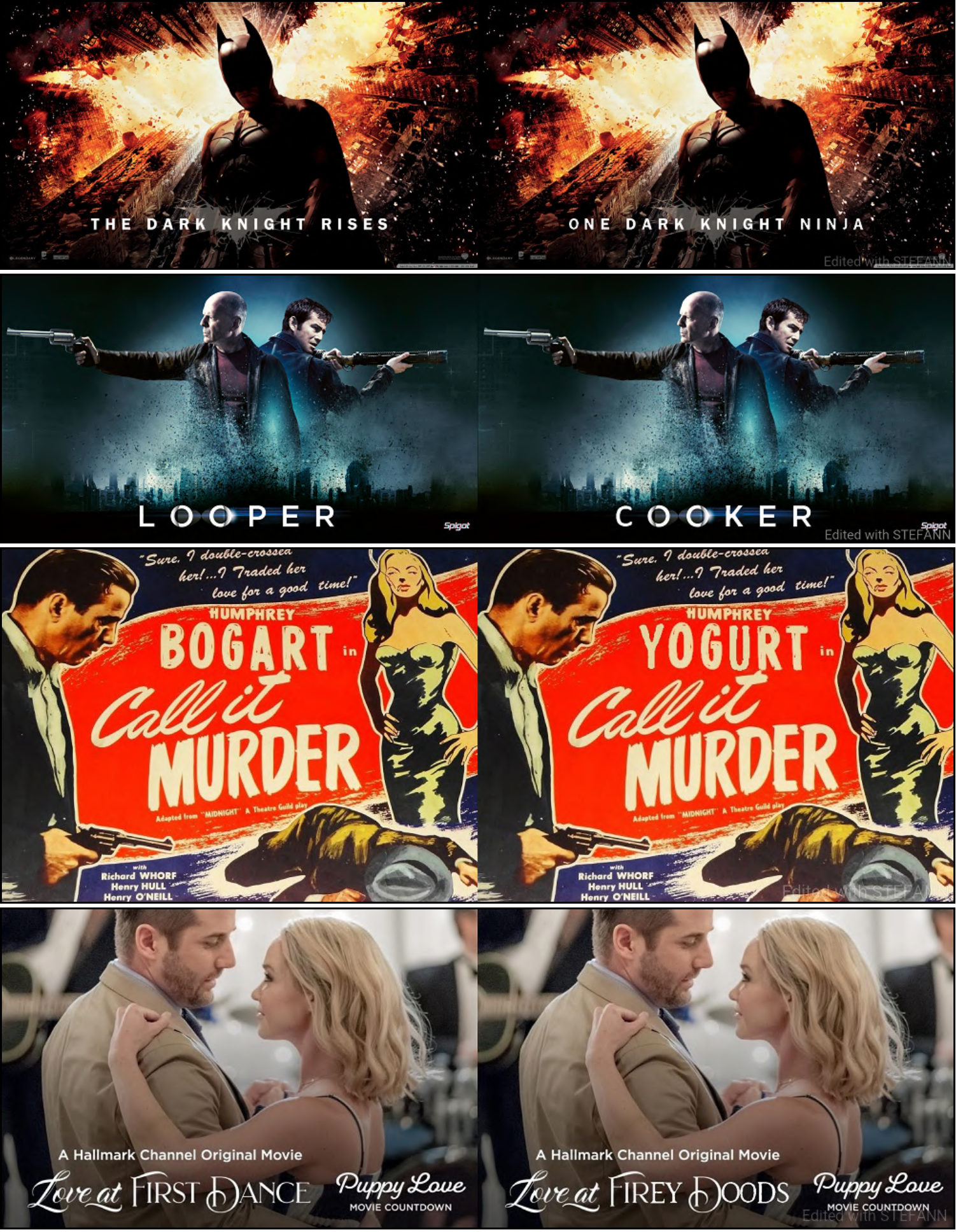}
\caption{Editing texts in movie posters with STEFANN. \textbf{Left:} Original images. \textbf{Right:} Edited images. Text regions are intentionally left unmarked to show the visual coherence of edited texts with original texts without creating passive attention.}
\label{fig:fig2}
\end{figure*}

\begin{figure*}
\centering
\includegraphics[width=\textwidth]{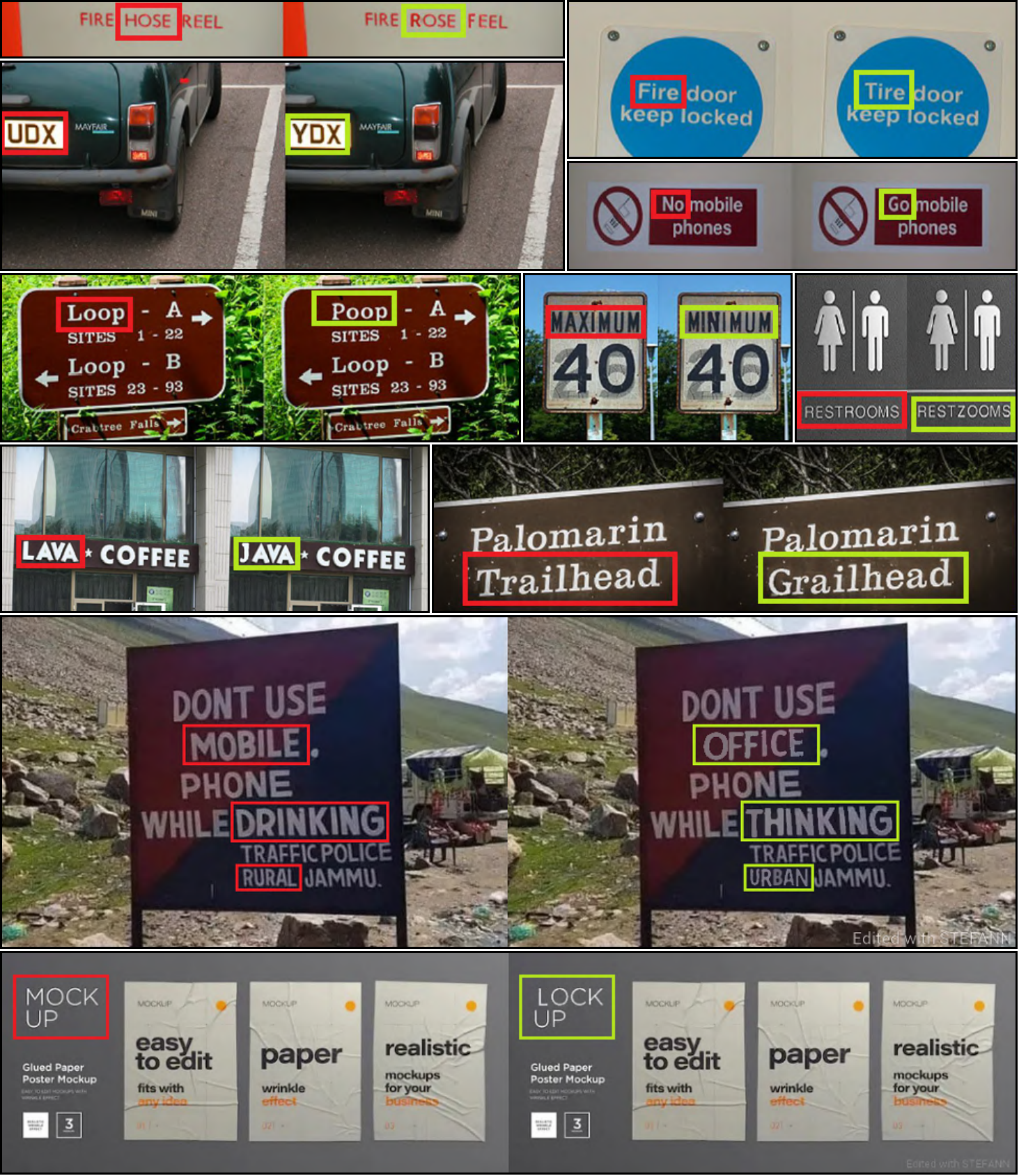}
\caption{Editing texts in scene images with STEFANN. \textbf{Left:} Original images with text regions marked in red. \textbf{Right:} Edited images with text regions marked in green.}
\label{fig:fig3}
\end{figure*}

\begin{figure*}
\centering
\includegraphics[width=\textwidth]{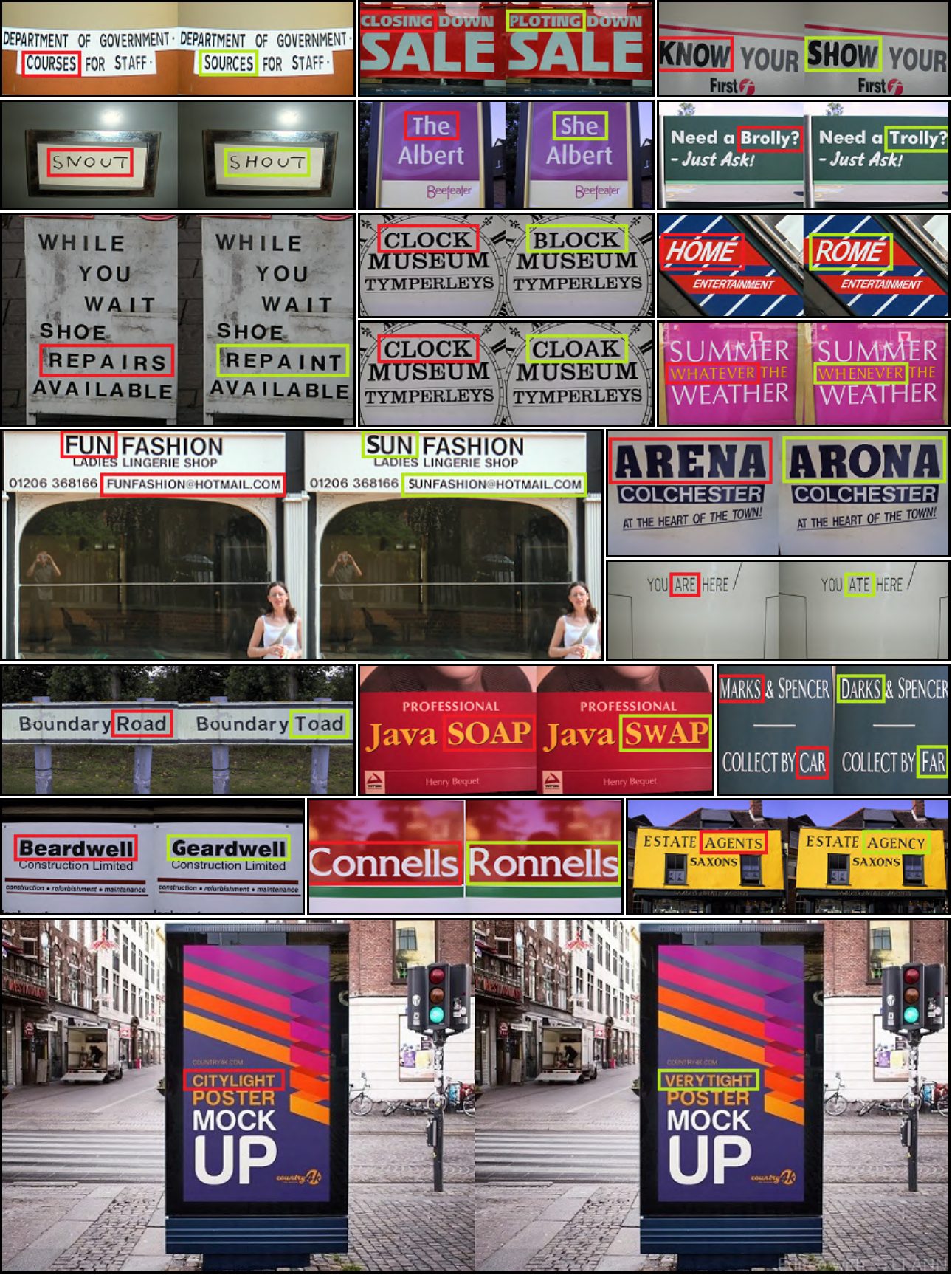}
\caption{Editing texts in scene images with STEFANN. \textbf{Left:} Original images with text regions marked in red. \textbf{Right:} Edited images with text regions marked in green.}
\label{fig:fig4}
\end{figure*}

\begin{figure*}
\centering
\includegraphics[width=\textwidth]{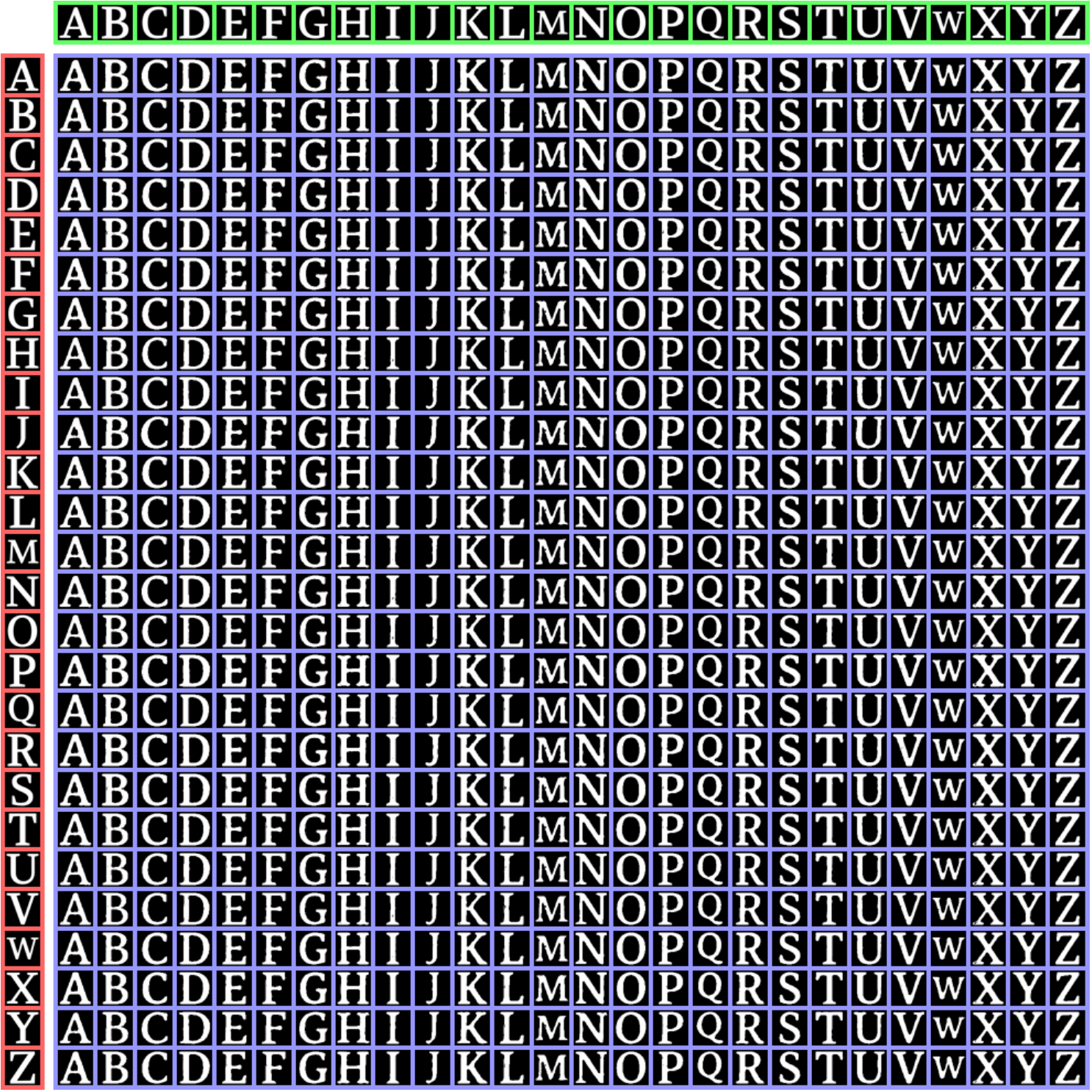}
\caption{Generation of all possible image pairs for a specific font with FANnet. In the first row, characters highlighted in green are the ground truth images. For each subsequent row, character highlighted in red is the image input (source) and characters highlighted in blue are the image outputs (targets) from FANnet, generated by varying the encoding input for each target character. This figure shows the structural consistency of FANnet on a specific font regardless of the source character.}
\label{fig:fig5}
\end{figure*}

\begin{figure*}
\centering
\includegraphics[width=\textwidth]{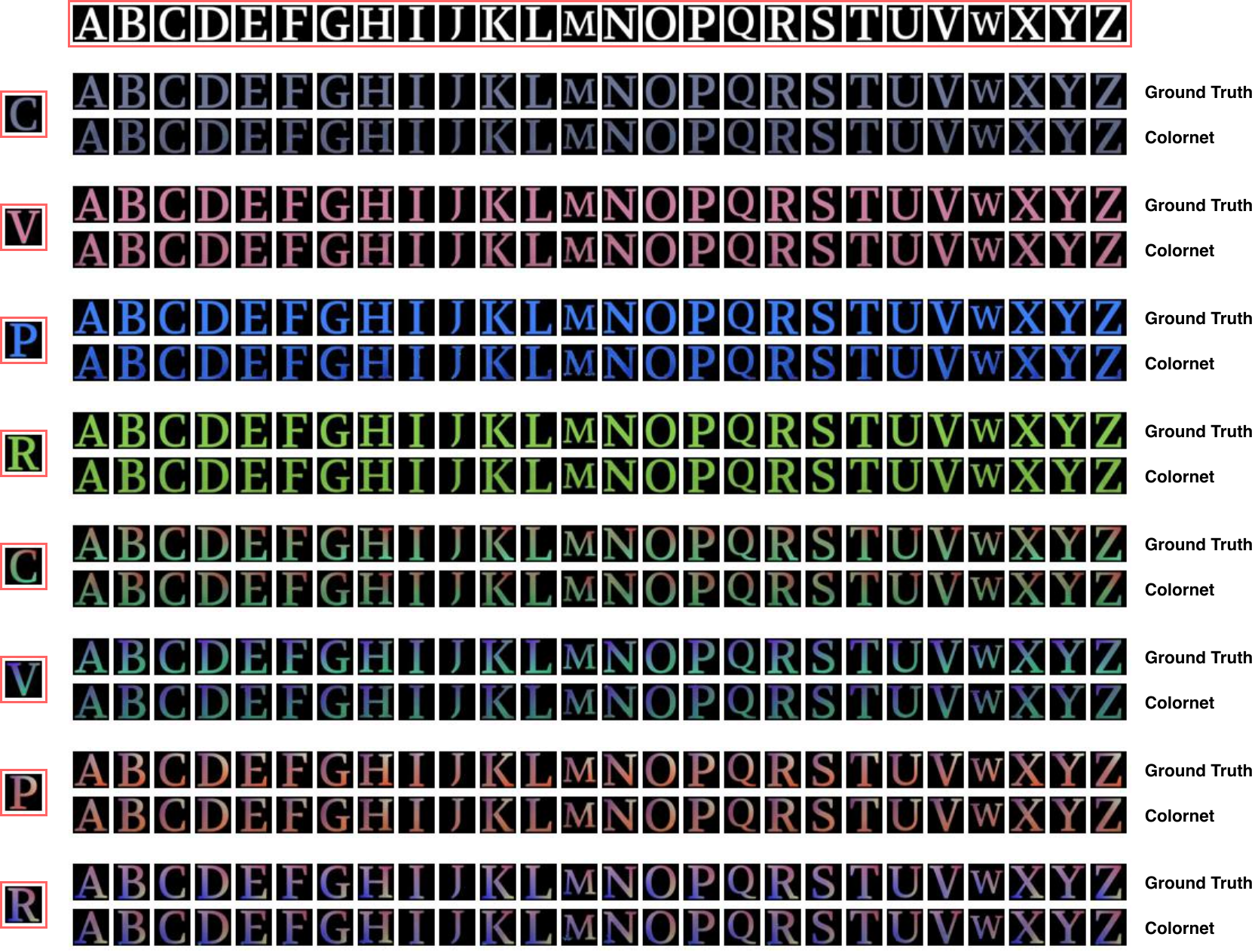}
\caption{Color transfer with Colornet. In the first row, characters highlighted in red are the binary target character images that need to be colorized. Each subsequent image block shows the colored source character image (highlighted in red), ground truth images for the colored target characters (top) and the color transferred images generated from Colornet (bottom). This figure shows the color consistency of Colornet for both solid and gradient colors.}
\label{fig:fig6}
\end{figure*}

% \end{document}

\end{document}